# Video Action Recognition Using spatio-temporal optical flow video frames

**Aytekin Nebisoy**
Kapital Bank
Consumer Experience Tribe
Baku, Azerbaijan
nabisoyaytekin@gmail.com

**Saber Malekzadeh**
Khazar University
Lumos Extension School of Data
Baku, Azerbaijan
saber.malekzadeh@sru.ac.ir

*Abstract*— Recognizing human actions based on videos has became one of the most popular areas of research in computer vision in recent years. This area has many applications such as surveillance, robotics, health care, video search and human-computer interaction. There are many problems associated with recognizing human actions in videos such as cluttered backgrounds, obstructions, viewpoints variation, execution speed and camera movement. A large number of methods have been proposed to solve the problems. This paper focus on spatial and temporal pattern recognition for the classification of videos using Deep Neural Networks. This model takes RGB images and Optical Flow as input data and outputs an action class number. The final recognition accuracy was about 94%.

## I. INTRODUCTION

Due to its importance in many applications such as intelligent surveillance, human computer interaction, assistance to senior citizens, or web-video search and access, the field of action recognition has grown dramatically recently. Many research trials have addressed action recognition as an obvious problem. [1] Different datasets are created to evaluate architectural variations. Different action recognition datasets were investigated in order to emphasize their ability to evaluate different models. In addition, a model is used for each dataset according to the content and format of the data it contains, the number of classes it covers, and the difficulties. On the other hand, another discovery is made for different architectures, showing the contribution of each to overcoming different action recognition problem challenges and the scientific explanation behind its consequences. Different datasets are created to evaluate architectural variations, and different action recognition datasets are explored to highlight their ability to create different models. On the other hand, another study of different architectures is being conducted, showing the contribution of each to solving different problems of recognition of actions and the scientific explanation underlying. [2] First of all, we need datasets according to the model. Datasets for video action recognition are defined by a set of qualities: source, pre-processing, point-of-view, number of videos, length of each video, number of action classes, classes per video (single or multi-label), and purpose. A myriad of datasets has been crafted and curated to span this spectrum of qualities. The vast majority of these datasets focus entirely on human actions due to their relevance in all aspects of everyday life. UCF101 [3] and HMDB51 [4] with 13,000 and 7,000 videos respectively, quickly became the main criteria for recognizing human actions. Thumos [5] and ActivityNet [6] pursued similar goals but did not achieve the same level of popularity.

In this paper, Bloopers (Video Blooper) dataset is used and on that dataset, deep learning is applied for action recognition. Proposed method is a spatio-temporal action recognition model with a combination of convolution and recurrent layers. Model takes RGB values, and Optical Flow preliminary estimated on the current video frame as input and outputs action class number.

## II. RELATED WORKS

**2D Approaches:** 2D approaches include several traditional networks: 2D Convolutional Neural Networks (C2D), Long-term Recurrent Convolutional Neural Networks (LRCN) referred to as CNN+LSTMs, Temporal Segment Networks (TSN), and Temporal Shift Modules (TSM). C2D models are derived directly from the image recognition area. With C2D, extracted frames from the video are used as input to 2D ConvNet. After several convolution and pooling layers, they are fed into one or more fully connected layers, which produce a SoftMax function output prediction over the dataset classes. With TSN, a video is segmented along its temporal (frame sequence) dimension and a frame is extracted from each segment for input to 2D ConvNet that share weights. The predictions from each segment are averaged before the SoftMax output layer. Variants and additions to TSN include Temporal Relations Networks (TRN), which performs multi-scale relations. In LRCN, the video is segmented and the frame is extracted from each segment. Those frames feed into 2D-ConvNets. However, the ConvNet outputs are used as inputs to a Long-Short Term Memory (LSTM) network before SoftMax predictions. [7]

**Two-stream networks** Since understanding the video intuitively requires motion information, finding a suitable way to describe the temporal relationship between frames is necessary to improving the performance of CNN-based video action recognition. Optical flow is an efficient motion representation to describe object/scene movement. More precisely, it is the pattern of clear motion of objects, surfaces, and edges in a visual scene caused by the relative motion between an observer and a scene. In Figure 1, we show visualization example of optical flow. As we can see, optical flow is able to define the motion pattern of each action accurately. The advantage of using an optical stream is that it provides orthogonal information by comparison with an RGB image. For example, the two images in Figure 2 have cluttered backgrounds. As shown in the figure optical flow can effectively remove the nonmoving background and result in a

simpler learning problem compared to using the original RGB images as input [8]

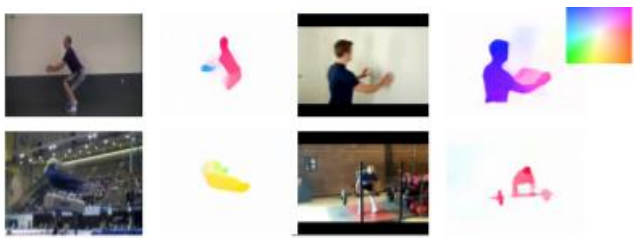

Figure 2. Visualizations of optical flow

**Two-stream fusion:** Because of there are two streams in a two-stream network, there will be a stage where it is necessary to combine the results from both networks to get the final predict. This stage is commonly referred to as the spatial-temporal fusion stage.

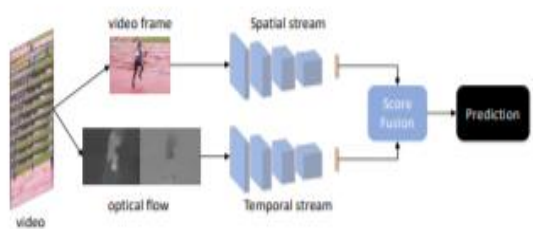

Figure 3. Two-stream Networks

**Recurrent neural networks**: Because of a video is mainly a temporal sequence, researchers have explored Recurrent Neural Networks (RNNs) for temporal modeling inside a video, particularly the usage of Long Short-Term Memory (LSTM) [8]

**Segment-based methods:** As shown in Figure 4, TSN first splits a whole video into several segments, where the segments distribute equally along the temporal dimension. Then TSN randomly selects a one video frame within each segment and forwards them through the network. In this case, the network shares weights for input frames from all the segments. Finally, a segmental concurrence were performed to aggregate information from the sampled video frames [8].

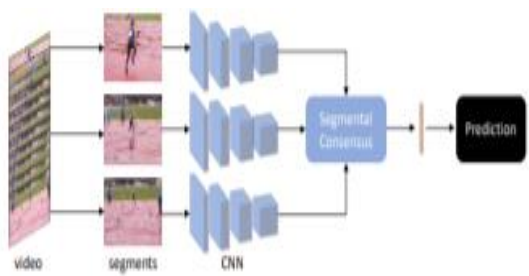

Figure 4. Temporal Segment Networks

**Multi-stream networks**: Two-stream networks are successful because contain two of the most important properties of a video which are appearance and motion information. . However, other factors can help video action recognition as more accurately, such as pose, object, audio and depth, etc. Skeleton information is closely related to human action. We can recognize most actions without scene context by just looking at a pose (skeleton) image. Although there is previous, work on using, pose for action recognition [9]. P-CNN was one of the first deep learning methods that successfully used pose to improve video action recognition. P-CNN has proposed to aggregate movement and appearance information along the traces of human body parts in a spirit similar to trajectory pooling [10] and has expanded this pipeline to a chained multi-stream framework that computes and integrates appearance, movement and posture.

**3D Approaches:** 3D Convolutional Neural Networks (C3D) were designed as the 3D analogy of 2D ConvNets. However, due to the long-term dependence of actions, C3D models are often less successful at recognizing actions than their two-dimensional counterparts at recognizing objects. To try to eliminate the gap between 2D and 3D models, Inflated 3D (I3D) models were created by "inflating" pretrained 2D kernels into 3D kernels. This allows I3D models to benefit from pretraining on 2D image datasets like ImageNet. Both C3D and I3D use either the whole video or a selected portion (for example, 16, 21, or 32 frames) as input to 3D-ConvNet. The 3D-ConvNet output is fed to the classification network before outputting SoftMax predicts similar to the C2D approach. [11] [12] [13]. The third trend focused on computational effectiveness to scale to even larger datasets so that they could be adopted in real applications. Examples include Hidden TSN [14], TSM [15], X3D [16], TVN [17] etc.

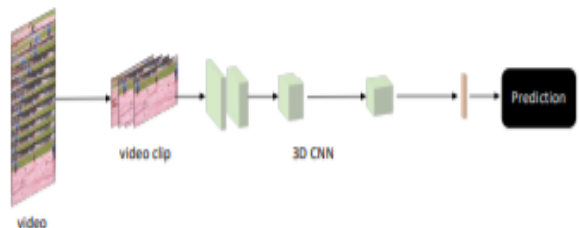

Figure 5. I3D

## III. DATASET

In this paper, Blooper dataset used for action classification. Bloopers are mistakes that a person makes on the screen. Monologues are the perfect place to spot the blunders of fixed conditions. The monologues involve one person on the screen and a fixed camera position. In this dataset, short videos have collected from YouTube and split this video clips into blooper and no blooper videos. There are 600 videos have time length are between 1 and 3 seconds per video clip. They split train, test and validation with two categories: zero Blooper and one No blooper. The videos are multilingual. Both genders are represented. [18]

## IV. METHOD

Classification of actions is performed on bloopers dataset for classify if there is blooper in video. The video frames are resized to 128×128 pixels and their Optical Flow is computed.

### A. Optical Flow

Optical flow is the motion of objects between consistent frames of sequence, caused by the relative movement between the object and camera. The problem of optical flow may be expressed as [19]:

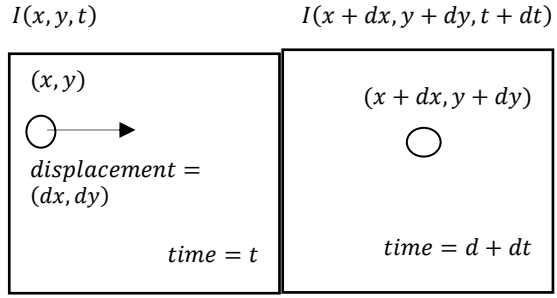

where between consistent frames, we can indicate the image intensity *(I)* as a function of space *(x,y)* and time *(t)*. In other words, if we take the first image *I(x,y, t)* and move its pixels by *(dx,dy)* over t time, we get the new image.

$$I(x + dx, y + dy, t + dt) \quad (1)$$

First, we assume that pixel intensities of an object are constant between consecutive frames. [19]

$$I(x, y, t) = I(x + dx, y + dy, t + dt) \quad (2)$$

Then, we take the Taylor Series Approximation of the RHS and remove general terms.

$$I(x + dx, y + dy, t + dt) = I(x, y, t) + \frac{\partial I}{\partial x}dx + \frac{\partial I}{\partial y}dy + \frac{\partial I}{\partial t}dt + \cdots \quad (3)$$

Third, we divide by dtdt to derive the optical flow equation:

$$\frac{\partial I}{\partial x}u + \frac{\partial I}{\partial y}v + \frac{\partial I}{\partial t} = 0 \quad (4)$$

$\frac{dI}{dx}, \frac{dI}{dy}$ and $\frac{dI}{dt}$ are the image gradients along the vertical axis, the horizontal axis, and time. Hence, we complete with the problem of optical flow, that is, solving u(dx/dt) and v(dy/dt) to define movement over time. We know that for u and v we cannot directly solve the optical flow equation. Because there is two unknown variables in one equation. We will implement some methods.

Dense optical flow try to compute the optical flow vector for all pixels of each frame. Although such computation may be slower, it gives a more accurate result and a denser result appropriate for applications such as learning structure from motion and video segmentation. [19]

A dense optical flow can be seen as a set of movement vector fields dt between the pairs of consistent frames t and t + 1. By dt(u, v) we denote the movement vector at the point (u, v) in frame t, which moves the point to the corresponding point in the following frame t + 1. The horizontal and vertical components of the dxt and dyt vector field, can be seen as image channels, well adapted to recognition using a convolutional network. To show the movement across a sequence of frames, we stack the flow channels dx,yt of L consistent frames to create a total of 2L input channels. Let height of a video be h and width be w [19]

$$I_\tau(u, v, 2k - 1) = d^x_{\tau+k-1}(u, v), \quad (5)$$

$$I_\tau(u, v, 2k) = d^y_{\tau+k-1}(u, v), \quad u = [1; w], v = [1; h], k = [1; L] \quad (6)$$

There are different implementations of dense optical flow. We will be using the Farneback method, one of the most popular implementations [20]

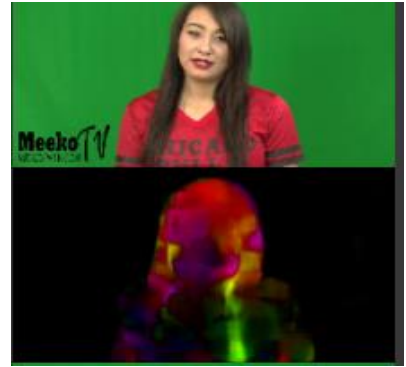

Figure 6. Dense Optical Flow with Farneback method

Gunnar Farneback proposed an effective method to calculate the motion of interesting features by comparing two consistent frames in his paper Two-Frame Motion Estimation Based on Polynomial Expansion [20].

First, the method guesses the windows of image frames by quadratic polynomials through polynomial expansion transform. Second, by observing how the polynomial transforms under translation (motion), a method to evaluate displacement areas from polynomial expansion coefficients is defined. After a series of clarifications, dense optical flow will computed. [20]

### B. Architecture

Our model is constituted of five 2D convolutional layers with 20, 30, 40, 50, 32 filter response maps, followed by a 3D Convolution layer with 50 filter and GRU layers and fully connected layer of size 200.

Unlikely LSTM, GRU tries to implement fewer gates. This method helps to lower down the computational cost.

In Gated Recurrent Units, an output gate controls the proportion of information that will be passed to the next hidden state. Furthermore, we have an input gate that controls information flow from current input. [21]

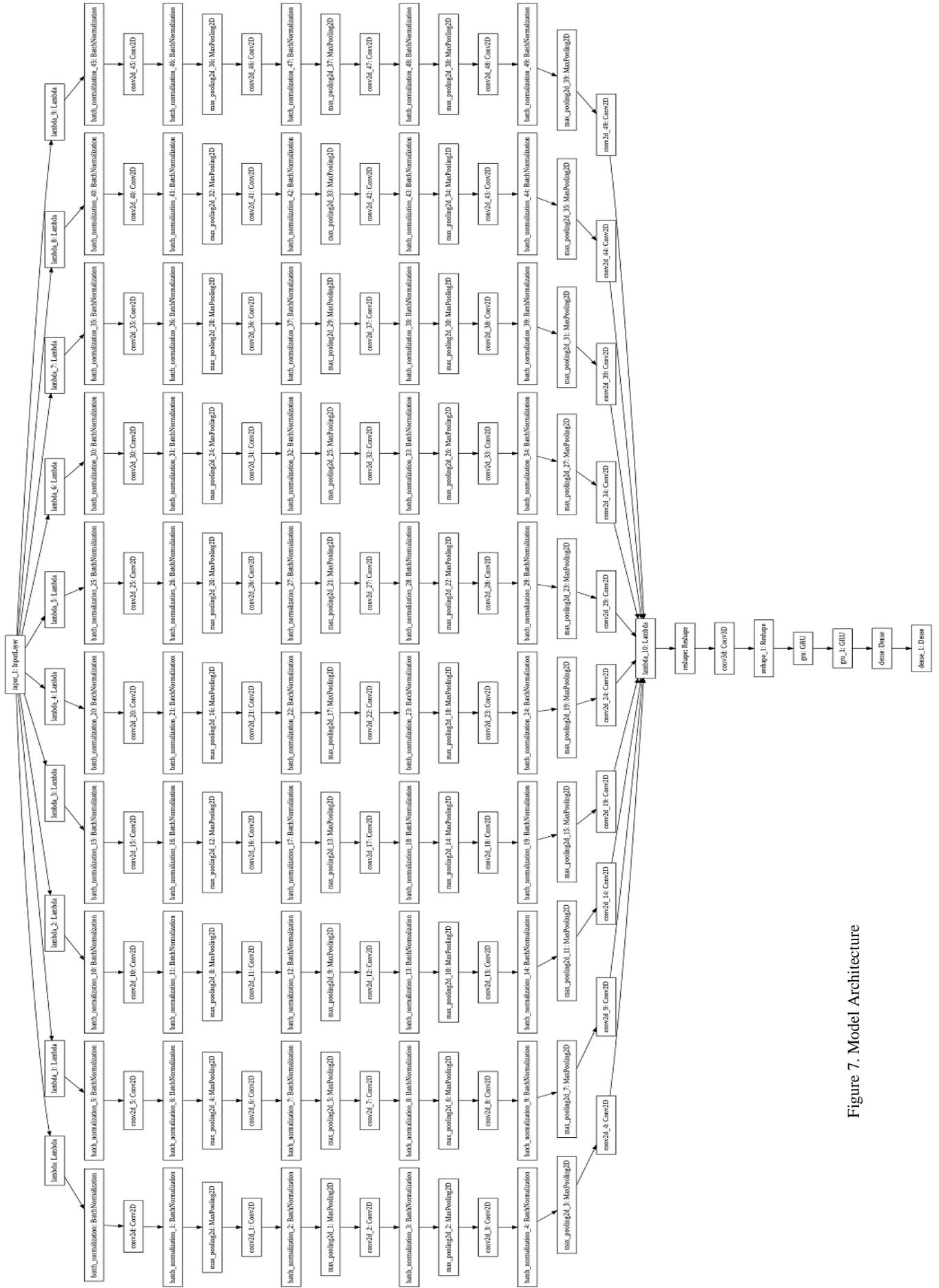

Figure 7. Model Architecture

Unlike RNN in GRU, forget gates are not used. [21]

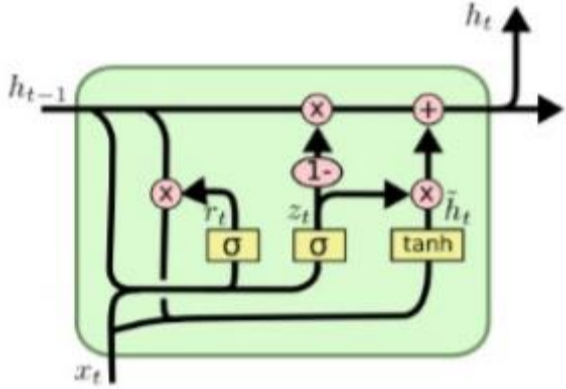

Figure 8. An illustration of GRU

$$zt = (Wz \cdot [htI, xt])$$

$$rt = (Wr \cdot [htI, xt])$$

$$tanh(W \cdot [rt \cdot htI, xt])$$

$$ht = (1\ zt) \leftarrow ht1 + zt \leftarrow \tilde{h}t \quad (7)$$

We have 414 videos in train set and 50 videos in test set. In train set, each video has 10 frame. Optical Flow computed these videos and split train test set also to take an input.

In this model before training model, optical flow and frame images are concatenated with frame size (20, 128, 128, 3). In this model, optimization method is Stochastic Gradient Descent.

Stochastic gradient descent is an optimization method for unconstrained optimization problems. Unlike Batch Gradient Descent, SGD guesses the true gradient of E (w, b) by considering a one training example at a time. [22] The SGDClassifier class implements the first order SGD training routine. The algorithm iterates over the training examples and for each example the model parameters are updated according to the update rule given by.

$$w \leftarrow w - \eta [\alpha \frac{\partial R(w)}{\partial w} + \frac{\partial L(w^T x_i + b, y_i)}{\partial w}] \quad (8)$$

Where η is the learning rate, which controls the size of step in the parameter space. Without regularization the intercept b is similarly updated .The learning rate η can be either constant or gradually regressing. The default learning rate schedule (learning rate='optimal') for classification, is given by.

$$\eta^{(t)} = \frac{1}{\alpha(t_0 + t)} \quad (9)$$

where t is the time step (there are a total of n samples * n iteration time steps), t0 is determined based on a heuristic proposed by Léon Bottou such that the expected initial updates can be compared with the expected size of the weights [22] As loss function used Categorical Cross Entropy [23]. Categorical Cross Entropy loss function is derived from the Log SoftMax and Negative Log Likelihood functions and it is used in classification problems with multiple classes. It is simple and fast, as well as proven strong in many cases. The purpose of this function is to carry the elements of the same class into an N-dimensional space, within certain regions. Cross Entropy loss function is defined by equation:

$$l(\bar{y}i, yi) = yi ln(\bar{y}i) \quad (10)$$

## V. CONCLUSION

In this paper, optical flow used for temporal pattern recognition and frame images for spatial pattern recognition. Frames of videos are entered to convolution layers and appended to a list for joining in new axis and pass into GRU to recognize spatio-temporal patterns. Then passed to Dense layer with neuron number 200.

To find result, used K-Fold Cross-Validation with k= 5. Results of each fold is showing below:

| Dataset | | | | |
|---|---|---|---|---|
| Folds | Training set | | Test set | Accuracy |
| Fold#1 | Test | Train | | 77.17% |
| Fold#2 | Train | Test | Train | 98.91% |
| Fold#3 | Train | Test | Train | 100% |
| Fold#4 | Train | | Test | 100% |
| Fold#5 | Train | | Test | 93.48% |

Table 1. 5-Fold Cross-Validation result

Average of accuracy result is 93.91%. In addition, Confusion Matrix is showing below:

| Classes | 1 | | 2 | | 3 | | 4 | | 5 | |
|---|---|---|---|---|---|---|---|---|---|---|
| Non-blooper | 43 | 3 | 45 | 1 | 46 | 0 | 46 | 0 | 40 | 6 |
| Blooper | 18 | 28 | 0 | 46 | 0 | 46 | 0 | 46 | 10 | 36 |

Table 2. Confusion Matrix

| Type | Accuracy |
|---|---|
| C2D [24] | 0.919 |
| LRCN [25] | 0.883 |
| C3D [26] | 0.842 |
| I3D [27] | 0.911 |
| AytekNet | 0.9391 |

Table1. Comparing Accuracy